# Effects of Different Attention Mechanisms Applied on 3D Models in Video Classification

Mohammad Rasras[1], Iuliana Marin[2][0000−0002−7508−1429], Șerban Radu[3], and Irina Mocanu[3][0000-0001-5176-9344]

[1] National University of Science and Technology POLITEHNICA Bucharest, Doctoral School of Automatic Control and Computers, Splaiul Independenței 313, 060042, Bucharest, Romania
`mohammad.rasras@stud.fils.upb.ro`
[2] National University of Science and Technology POLITEHNICA Bucharest, Faculty of Engineering in Foreign Languages, Splaiul Independenței 313, 060042, Bucharest, Romania
`iuliana.marin@upb.ro`
[3] National University of Science and Technology POLITEHNICA Bucharest, Faculty of Automatic Control and Computers, Splaiul Independenței 313, 060042, Bucharest, Romania
`{serban.radu,irina.mocanu}@upb.ro`

**Abstract.** Human action recognition has become an important research focus in computer vision due to the wide range of applications where it is used. 3D Resnet-based CNN models, particularly MC3, R3D, and R(2+1)D, have different convolutional filters to extract spatiotemporal features. This paper investigates the impact of reducing the captured knowledge from temporal data, while increasing the resolution of the frames. To establish this experiment, we created similar designs to the three originals, but with a dropout layer added before the final classifier. Secondly, we then developed ten new versions for each one of these three designs. The variants include special attention blocks within their architecture, such as convolutional block attention module (CBAM), temporal convolution networks (TCN), in addition to multi-headed and channel attention mechanisms. The purpose behind that is to observe the extent of the influence each of these blocks has on performance for the restricted-temporal models. The results of testing all the models on UCF101 have shown accuracy of 88.98% for the variant with multiheaded attention added to the modified R(2+1)D. This paper concludes the significance of missing temporal features in the performance of the newly created increased resolution models. The variants had different behavior on class-level accuracy, despite the similarity of their enhancements to the overall performance.

## 1   Introduction

One of the most significant areas in the computer vision discipline is human action recognition (HAR), which depends on data gathered by sensors, such as cameras [1]. This field has been involved in various domains, for instance in video surveillance [2, 3], healthcare assistance [4], sports analytics [5], and entertainment [6]. Moreover, deep learning algorithms, including Convolutional (CNN) and Recurrent (RNN) Neural Networks, have played a role in the development of the computer vision field in the last decade [7]. For instance, image classification has benefited not only from the birth of large datasets, such as ImageNet [8], but also from the deeper designs of 2D CNNs, including VGG-16 [9], AlexNet [10], Inception [11], ResNet [12], and DenseNet [13]. Those new models have shown capabilities in extracting spatial features from images and, consequently, a remarkable improvement in performance.

Motivated by the success of 2D architectures on still images, researchers started to investigate the ability to apply these frameworks on video classification tasks. However, the results were lower compared to when applied on still images. The performance gap between image and video classifiers is due to the nature of videos, which are simply sequences of frames that associate motion with them. As a result, this made scientists think of other techniques to overcome the weaker accuracy obtained on clips.

Prior to the advent of Transformer models, deep learning vision-based human action recognition relied on three major approaches, namely two-streams 2D CNN, RNN, and 3D CNN-based models. As its name indicates, the first technique consists of two branches. One is used to extract spatial features from RGB images, while the second operates on the temporal information obtained from optical flow data [14, 15]. RNNs, on the other hand, have a built-in memory that allows handling sequences of data, but struggled due to the vanishing grading descend, which was better managed in its gated Long-Short Term Memory (LSTM) variant. Generally, RNNs and their variants are not well-suited for extracting spatial features, and hence, they were embodied within models that combine them with 2D CNN frameworks to overcome this issue [16-18].

Expanding deep 2D classification models, such as Inception and Resnet [17, 19], into 3D ones was the main interest of researchers [20]. The output of these extended designs demonstrated the ability to extract both spatial and temporal information. In addition, the performance of these 3-dimensional setups can compete and even outperform, in some cases [21-23], both RNN and two-stream 2D CNN-based state-of-art methods. Moreover, the ease of implementation and the direct employment of 3D CNNs on RGB videos without the extra need to extract the optical flow for motion, contributed to making these architectures the most dominant approaches [24].

In this paper, we considered the three R3D, MC3, and R(2+1)D frameworks [25], as we first created three identical to the original architecture to be trained on UCF101 [26], except a dropout layer was added before the final network, which then served as backbones used for developing ten distinct variants. Each of these variants is distinguished by a specific attention unit. We setup our experiment to first assess the impact of reduced temporal and increased spatial features on these backbones' performance, and then we studied the improvement that occurred in each variant.

The paper is organized as: Chapter 2 outlines related work regarding HAR 3D CNN models, followed by Chapter 3, that describes the experimental setup comprising data preprocessing, training, and model configurations. Analysis and discussions obtained from the results of our experiment are presented in Chapter 4. The last chapter is related to conclusions and future work.

## 2       Related Work

State-of-the-art research evolves around different methods, such as two-streams, RNN, and transformers, in addition to 3D-based models covered in this section.

One of the earliest HAR 3D CNN-based designs was done by [19], their named C3D novel model has eight convolutional layers with a common kernel size of 3 on all dimensions. The outcome of this work proved competitive capability of capturing spatiotemporal data compared to current state-of-the-art models. Afterwards, some researchers, such as [27], transferred the concept of extending current powerful deep 2D image classifiers, particularly DenseNet, into 3D setups. The results demonstrated higher accuracy on benchmark datasets compared to other HAR models.

The success of expanding deep 2D frameworks into 3D kernelled ones continued, as how authors in [28] performed on Inception as a backbone model. Because of the high accuracy achieved on RGB input, authors also considered utilizing it in a two-stream setup that employs the optical flow in it. Inflating 2D models into 3D ones is usually accompanied by a large amount of data. This makes the training of these frameworks computationally expensive. Therefore, researchers responded to this issue by representing the 3D kernelled networks differently, i.e. splitting them into a combination of temporal and spatial parts. This technique prevailed in many 3D-based configurations. One of the state arts works that resembles this concept was carried out by authors in [25], who built a mixed setup of the 3D and 2D CNNs based on the original 2D ResNet. The first two layers employ the 3D CNNs, whereas the rest use the 2D ones.

The concept of factorizing the 3D network showed not only an improvement in efficiency, but also an increase of overall performance compared to standard 3D Resnet. Thus, other mainstream research continued to apply this mechanism. Another design proposed in the paper [29] decomposes the three-dimensional convolutions in the model into sequentially separate 2D and 1D to capture spatial and motion, respectively. Adoption of this mechanism yielded an increase in accuracy and a drop in a number of parameters compared to the original three-dimensional one. Impressed by their outcomes, the authors replicated the model as part of a two-stream configuration.

In another way of focusing on separating spatial and temporal extraction methods, authors in [30] introduced a novel model that uses two paths-mechanism for handling this compound data. The slow path in this model operated on high-resolution frames, while the fast one handled the temporal dynamics. The contrast in this configuration demonstrated high performance in video classification tasks.

Authors in [31] built a two-stream 3D model. As 3D architecture can capture both spatial and temporal data, the authors implemented a second 3D stream that operates on optical flow. This second added path was used in training to influence the first RGB stream by transferring knowledge, which is a setup also known as the teacher-student framework. In this case, the teacher stream works on optical flow, representing motion, and then distills its learned extracted capabilities on temporal dynamics to the student path that handles the RGB input. Other state-of-the-art methods, such as in [32], addressed the utilization of 3D models by introducing a block that groups similar temporal features and consequently reduces the number of computations. Designers of this work stated a 33-53% reduction in computation cost, while maintaining the same or higher performance. Another 2D backbone was expanded to a 3D video classifier, as [33] did on VGG-16. The authors of this work combined the strength of image classification achieved by the VGG-16 network with the LSTM capability for handling sequences. The designed model of this work is suitable to run in a real-time setup.

In this paper, we adopt the three models proposed by [25], namely the R(2+1)D, R3D, and MC3, as we mentioned previously. The main reason for our choice, rather than their high performance, is that although these three frameworks share the same 18 layers of depth and are based on the ResNet 2D framework, they differ in the way they implement their convolutional filters which we will address in the coming sections.

## 3  Experimental Setup

We started our experiment by implementing different mechanisms compared to the models proposed by [25]. As we initially applied a functionality that checks the number of frames in a clip not to exceed a limit of 48. If the video contains fewer than that, all of them will be extracted, otherwise, a uniform subsampling is applied to extract exactly the maximum allowed number. The purpose of this configuration is that videos in UCF101 vary in length, and we wanted to have consistent temporal content out of them, in other words, we intentionally reduced the temporal knowledge in longer videos.

In the first part of the experiment, we modified the original MC3, R3D, and R(2+1)D [25] to create backbones for each, by adding a dropout layer, with a value of 0.4, for each one of them before the fully connected layer (fc), to increase the generalization of models. From now on in this article, we refer to those backbones as M-MC3, M-R3D, and M-R(2+1)D, based on which we later propose ten variants that include different attention mechanisms.

The UCF dataset contains 101 actions, distributed in five categories related to the actions that occur either with objects, including musical instruments, or with another individual, in addition to sports. Figure 1 demonstrates some extracted frames for random actions.

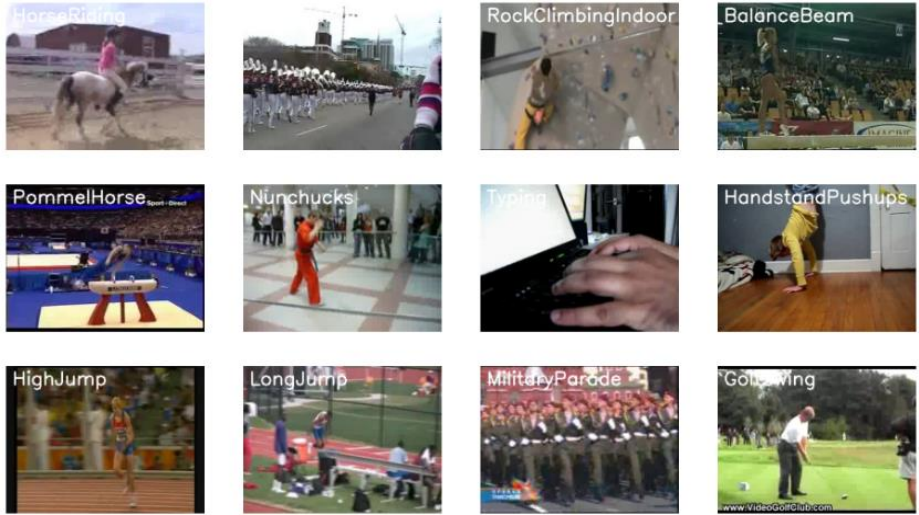

**Fig. 1.** Samples of extracted frames extracted from twelve actions of UCF101 dataset.

Apply eye makeup, lipstick, and yo-yo are actions in the human-object category, examples of body-motions are handstand pushups, handstand walking, and walking with dog. The human-human interaction includes class labels, such as head massage, band marching, and haircut. Playing musical instruments has activities, like playing cello, daf, and dhol. The sport's category contains actions like parallel bars, long jump, and cliff diving. The dataset's clips have 25 frames per second and a resolution of 320 x 240. The total number of videos is 13,320, with an average of 7.21 seconds in length.

### 3.1  Data Preprocessing

After the videos were preprocessed initially to have a maximum of 48 frames, as described previously, they are passed afterwards to the data augmentation pipeline. Here, another functionality ensures the clip is at least 32 frames long. If they are shorter, the last frame is then repeated to pad the sequence until it reaches that minimum value. A temporal subsampling technique then operates to select 16 consecutive frames with random starting points, to create temporal variability that enhances the model's training process of learning.

Spatial transformation is built differently compared to the original setup, as it first resizes each frame to 256 x 256 pixels, and then crops it at the center to have a resolution of 224 x 224, where the author's work in [25] had 112 x 112 instead. Additionally, color jittering and horizontal flipping were then applied. Similarly, the same methodology of temporal augmentation was implemented in the testing split, except that the temporal subsampling generates three clips for each video processed, all with different random starting points to enhance evaluation results by averaging the three values.

There are two main differences between our data preprocessing approaches and the original work, which are related to the temporal and spatial features passing to the model. Our goal, by this contrast, is to study the impact of the reduced temporal and the increased spatial features on the overall model's performance. We used UCF101 as a benchmark dataset for our tests and split 1 was adopted for the training and evaluation.

### 3.2  Models' Configuration

Based on the original MC3, R3D, and R(2+1)D models, which share the same number of layers, and were pretrained on the Kinetics-400 [34] dataset, we started our experiments by finetuning on UCF101 the new M-MC3, M-R3D, and M-R(2+1)D models that differ from originals by only adding the dropout layer before fc, as mentioned above. Figure 2 demonstrates the architecture of the original frameworks.

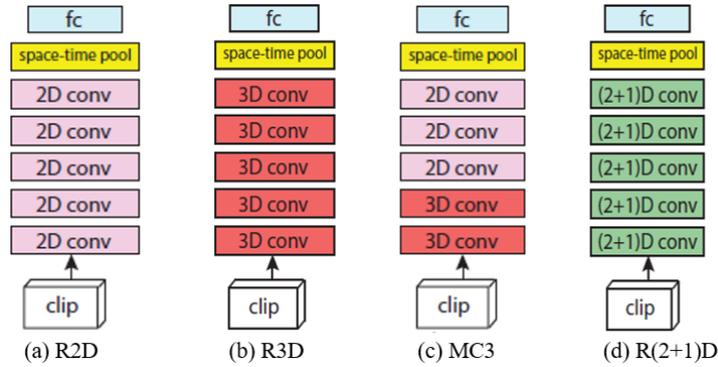

**Fig. 2.** The architectures of (a) 2D ResNet; (b) R3D; (c) MC3; and (d) R(2+1)D models. [18]

The ten new variants that we proposed based on the modified backbones contain attention blocks, namely temporal attention [33] that aids in improving the temporal feature representation by focusing on relevant information within the sequence of frames. Squeeze and excitation mechanism (SE) [35] that focuses on certain channels used in determining features by effectively enhancing most important feature maps, while ignoring irrelevant ones. The Convolutional Block Attention Module (CBAM) [36], which is a type that utilizes both channel and spatial types of attention, by allowing the network to refine features in both channel and spatial dimensions. Temporal Convolutional Network (TCN) [37], which is powered with a single temporal dimension 1D CNN, helps in modelling long dependencies, due to large receptive fields they have. Figure 3 illustrates the ten created variants.

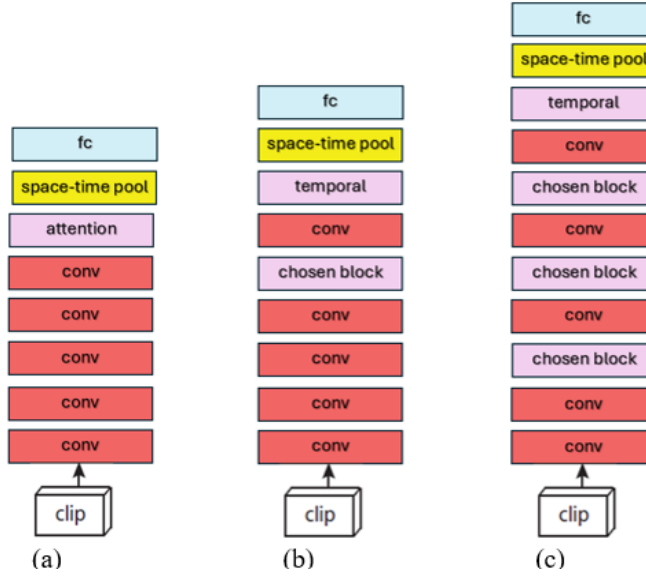

**Fig. 3.** The ten proposed architectures all incorporating a dropout layer before the fully connected (fc) layer. (a) Variants where the attention block is either spatial or temporal. (b) Variants incorporating specific attention mechanisms, including SE, temporal, a combination of SE and multi-head attention, TCN, and CBAM, applied after layer 3. (c) Variants where the same attention mechanisms (excluding the spatial attention from (a)) are applied after layers 1, 2, and 3.

The ten new variations, that all contain the defined dropout layer before the fc one, are named after the block of attention, and their placement, as follows:

- FC-Spatial: adds multi-headed attention on spatial features before final pooling layer.
- FC-Temporal: employs multi-headed temporal attention before the final pooling layer.
- 3-SE: is a variant of FC-Temporal with an SE block after layer 3.
- 3-Temporal: is another variant of FC-Temporal, but it differs in adding a Temporal multi-headed attention layer before and after layer 3.
- 3-Both: employs Temporal multi-headed and SE units after layer 3, while following the same strategy in keeping the final temporal block before the final pooling.
- 3-CBAM: keeps the final temporal layer but implements CBAM after layer 3.
- 3-TCN: is a model like the previous one, except that it adopts the TCN layer instead.
- All-SE: keeps the final temporal layer and adds SE units after layers 1, 2, 3.
- All-Temporal: is the same as All-SE, but instead of SE blocks, it applies on the same placements, multi-headed attention that operates on temporal features.
- All-Together: is based on All-SE and All-Temporal, but uses both SE and multiheaded temporal attentions, instead of separate ones on same locations within the framework.

### 3.3   Training

In our training process, we employed the Stochastic Gradient Descent (SGD) optimizer with a momentum of 0.9. The learning rate was initially set to 0.001 and a scheduler that drops it by 0.1 factor every 15 epoch was also added. We also considered a weight decay with a value of 0.0005 to enhance the regularization while training the models. Moreover, we applied an early stopping mechanism to avoid overfitting, which also aided in minimizing the training time.

## 4      Analysis and Discussion

We performed our experiments on the ten variants described in section 3.3. Tables 1, 2, and 3 demonstrate the models, and the epoch at which they stopped training when early stopping was activated, params represent the number of learnable parameters expressed in millions, along with top-1 and top-5 accuracies. Additionally, the tables include the number of classes that show an increase on their original (M-MC3, M-R3D, and M-R(2+1)D) accuracy, represented by the no. inc., in the sixth column.

The highest increase and lowest decrease columns are related to the class that experienced the greatest accuracy increase and decrease among the 101 actions when tested by the corresponding variant. The table's final column (least class / acc) is the measure of the lowest accuracy achieved and its associated class label.

**Table 1.** M-MC3 variants analysis.

| Variants | Epoch | Params | top-1 | top-5 | No. inc. | Highest increase | Lowest increase | Least class / acc |
|---|---|---|---|---|---|---|---|---|
| M-MC3 | 28 | 11.54 | 81.21 | 95.88 | N/A | N/A | N/A | HighJump / 18.92 |
| FC-Spatial | 26 | 12.59 | 83.82 | 96.43 | 51 | Basketball / 34.28 | BreastStroke / -25.00 | HighJump / 29.73 |
| FC-Temporal | 25 | 12.59 | 83.4 | 96.46 | 49 | PommelHorse / 34.28 | BreastStroke / -35.71 | HighJump / 24.32 |
| 3-SE | 22 | 12.65 | 83.8 | 96.67 | 50 | WalkingWithDog / 41.67 | BreastStroke / -28.57 | PizzaTossing / 30.30 |
| 3-Temporal | 24 | 12.85 | 84.69 | 97.33 | 51 | WalkingWithDog / 44.45 | BreastStroke / -28.57 | HighJump / 40.54 |
| 3-Both | 24 | 12.92 | 84.54 | 96.51 | 54 | HighJump / 32.43 | Archery / -21.95 | Nunchucks / 34.29 |
| 3-CBAM | 23 | 12.60 | 85.62 | 96.67 | 61 | MoppingFloor / 29.41 | HammerThrow / -22.22 | CricketBowling / 41.67 |
| 3-TCN | 21 | 12.79 | 85.01 | 96.67 | 50 | PommelHorse / 48.57 | PlayingSitar / -22.73 | HandstandWalking / 35.29 |
| All-SE | 22 | 12.68 | 83.85 | 96.03 | 55 | HighJump / 32.43 | Kayaking / -19.44 | PizzaTossing / 30.30 |
| All-Temporal | 25 | 12.94 | 86.1 | 96.64 | 63 | HighJump / 35.13 | Haircut / -27.27 | PizzaTossing / 33.33 |
| All-Together | 22 | 13.02 | 84.88 | 96.59 | 54 | Basketball / 37.14 | HandstandPushups / -35.71 | HandstandWalking / 38.24 |

Table 1 shows that all variants created based on the M-MC3 have trained in a smaller number of iterations compared to the M-MC3 model, this implies that despite an increased number of parameters on the variants, they could train faster than the backbone. The top 1 accuracy ranged between 81.2 and 86.1% for the M-MC3 and All-Together models, respectively.

Among the variants that employed different blocks after layer 3 within its architecture, the 3-CBAM achieved the highest top 1 accuracy, followed by 3-TCN, where 3-SE had the lowest. Additionally, 3-Temporal recorded the highest top 5 accuracy among all variants, where 3-CBAM and 3-TCN share the same value with 96.67%.

Regarding the tests performed on M-MC3 and its based modifications, we found that all the variants have roughly increased half of the dataset classes' accuracies, for instance, the All-Temporal and 3-CBAM incremented the accuracy of 63, and 61 classes, respectively. Table 1 also indicates that the 3-TCN model has the greatest impact on class "PommelHorse", making it have the highest increase in accuracy.

Both "BreastStroke", and "HandstandPushups" were among all the 101 actions that were negatively affected by FC-Temporal and All-Together models, as they dropped 35.71% from their value on the original M-MC3 backbone.

The last column in Table 1 shows the lowest class accuracy for each variant, here the M-MC3 had the lowest accuracy, 18.92%, associated with "HighJump", the class remained the most misclassified in FC-Spatial, FC-Temporal, 3-temporal vases, despite an increase on its performance compared to the one recorded by the backbone. The 3-CBAM setup recorded the lowest accuracy associated with the "CricketBowling" class, with a value of 41.67%.

Table 2 illustrates that M-R(2+1)D based designs are all trained relatively within the same range of epochs. The minimum top 1 accuracy was obtained by the base model M-R(2+1)D, and the highest was reached by the 3-Temporal variant as 85.14 and 88.98%, respectively, the latter also recorded the first place among all designs for the highest top 5 value as well.

**Table 2.** M-R(2+1)D variants analysis.

| Variants | Epoch | Params | top-1 | top-5 | No. inc. | Highest increase | Lowest decrease | Least class / acc |
|---|---|---|---|---|---|---|---|---|
| M-R(2+1)D | 24 | 31.35 | 85.14 | 96.96 | N/A | N/A | N/A | PizzaTossing / 27.27 |
| FC-Spatial | 26 | 32.40 | 87.84 | 97.7 | 49 | BreastStroke / 57.15 | RockClimbingIndoor / -19.51 | PizzaTossing / 42.42 |
| FC-Temporal | 26 | 32.40 | 88.82 | 97.81 | 51 | BreastStroke / 50.00 | Shotput / -21.74 | Hammering / 45.45 |
| 3-SE | 25 | 32.46 | 87.02 | 97.75 | 45 | BreastStroke / 50.00 | PullUps / -21.43 | BrushingTeeth / 36.11 |
| 3-Temporal | 26 | 32.66 | 88.98 | 97.86 | 51 | JavelinThrow / 35.48 | FrontCrawl / -16.21 | Nunchucks / 51.43 |
| 3-Both | 22 | 32.73 | 87.63 | 97.65 | 51 | BreastStroke / 50.00 | PommelHorse / -22.86 | PizzaTossing / 39.39 |
| 3-CBAM | 22 | 32.41 | 88.55 | 97.57 | 53 | WalkingWithDog / 33.34 | HandstandPushups / -28.57 | PizzaTossing / 42.42 |
| 3-TCN | 22 | 32.60 | 88.21 | 97.38 | 52 | PizzaTossing / 42.43 | SalsaSpin / -18.61 | Nunchucks / 40.00 |
| All-SE | 23 | 32.49 | 86.52 | 97.52 | 51 | BreastStroke / 39.29 | Shotput / -32.61 | Hammering / 27.27 |
| All-Temporal | 25 | 32.75 | 87.89 | 97.62 | 44 | BreastStroke / 57.15 | Shotput / -36.96 | BrushingTeeth / 44.44 |
| All-Together | 23 | 32.83 | 88.9 | 97.54 | 53 | HighJump / 43.24 | BenchPress / -22.92 | Nunchucks / 48.57 |

Table 2, also shows the number of classes that had an increase in accuracy, among all variants, All-Temporal has influenced 44 actions to have an increment on the accuracy compared to backbone, although the variant distinguished "BreastStroke" class by making it the most to gain performance, in addition to making this value the highest increase to an action among the 11 testing models, with an increase of 57.15% to its original accuracy recorded by M-R(2+1)D.

Generally, the "BreastStroke" class was positively affected by M-R(2+1)D-base models. On the other hand, the action "Shotput" and "HandstandPushups" reduced their accuracies the most among all actions in the UCF101 dataset by testing All-Temporal and 3 CBAM.

The minimum class accuracy, according to Table 2, was of 27.27% for both "PizzaTossing" and "Hammering" activities, that were achieved by the original M-R(2+1)D and its base variant All-SE one, respectively. In addition, all the values of the lowest accuracies on classes were below 50%, except the "Nunchucks" one, which was obtained by testing the 3-Temporal setup.

The last ten variants in this part of our experiment are related to M-R3D and its variants. Like Tables 1 and 2, Table 3 illustrates the behavior of each of these models.

Based on the results from Table 3, the M-R3D based variant finished training in a range of 22-26 epochs. The values of the top-1 accuracy were between 79.99% and 84.44% for the original backbone and 3-SE models, respectively. For the variants with different blocks after layer 3, 3-SE achieved the highest top-1 accuracy with a score of 84.88%, followed by the 3-CBAM with 83.48%, while 3-TCN had the lowest value among them, with 82.16%. Nevertheless, All-Temporal had the greatest top-5 accuracy among the ten variants.

**Table 3.** M-R3D variants analysis.

| Variants | Epoch | Params | top-1 | top-5 | No. inc. | Highest increase | Lowest decrease | Least class / acc |
|---|---|---|---|---|---|---|---|---|
| M-R3D | 26 | 33.21 | 79.99 | 95.08 | N/A | N/A | N/A | PizzaTossing / 9.09 |
| FC-Spatial | 25 | 34.26 | 84.06 | 95.66 | 56 | TennisSwing / 38.78 | Shotput / -23.91 | PizzaTossing / 21.21 |
| FC-Temporal | 26 | 34.53 | 82.32 | 95.77 | 52 | BrushingTeeth / 41.67 | TaiChi / -21.43 | PizzaTossing / 30.30 |
| 3-SE | 24 | 34.33 | 84.88 | 96.46 | 57 | TennisSwing / 46.94 | BreastStroke / -25.00 | PizzaTossing / 15.15 |
| 3-Temporal | 26 | 34.53 | 82.32 | 95.77 | 52 | BrushingTeeth / 41.67 | TaiChi / -21.43 | PizzaTossing / 30.30 |
| 3-Both | 24 | 34.59 | 82.63 | 95.59 | 51 | MoppingFloor / 32.36 | BreastStroke / -21.43 | PizzaTossing / 21.21 |
| 3-CBAM | 22 | 34.27 | 83.48 | 96.27 | 60 | TennisSwing / 44.90 | BreastStroke / -35.72 | PizzaTossing / 30.30 |
| 3-TCN | 23 | 34.46 | 82.16 | 95.88 | 48 | BrushingTeeth / 36.12 | HandstandWalking / -29.41 | PizzaTossing / 12.12 |
| All-SE | 27 | 34.35 | 84.06 | 96.48 | 57 | JavelinThrow / 35.49 | HandstandPushups / -21.43 | HandstandWalking / 26.47 |
| All-Temporal | 25 | 34.61 | 84.11 | 96.67 | 60 | TennisSwing / 44.90 | BreastStroke / -46.43 | BreastStroke / 21.43 |
| All-Together | 23 | 34.70 | 83.45 | 95.93 | 57 | TennisSwing / 38.78 | HandstandWalking / -20.59 | PizzaTossing / 18.18 |

Table 3 also has shown that models 3-CBAM and All-Temporal enhanced performance in 60 classes. The 3-SE model has increased the accuracy to the highest value with 46.94% on the "TennisSwing" activity, followed by 44.90% for the same class tested by 3-CBAM and All-Temporal variants. The "BreastStroke" class has reduced its accuracy the most in the All-Temporal model's testing. The most frequently misclassified class is "PizzaTossing", that recorded the highest accuracy when testing the 3-Temporal and 3-CBAM variants, despite its original 9.09% value by M-R3D.

When studying these three tables, we noticed that the highest top-1 accuracy in each table was achieved by variants with implementations, for instance, 3-Temporal based on both M-MC3 and M-R(2+1)D recorded the highest value, whereas the case was for 3-SE based on backbone M-R3D. According to the three tables, the 3-CBAM variant, despite its backbone model, has influenced the greatest number of classes to improve their performance. The most increase to a class label among the tests on M-MC3 was for "PommelHorse", with an increment of 48.57%, which was achieved by the model 3-TCN. Differently, the case was for both variants, FC-Spatial and All-Temporal, based on M-R(2+1)D for the "BreastStroke" action.

The last backbone, M-R3D, also had a different variant that added the highest increase on performance to an activity, represented in 3-SE when detecting the "TennisSwing" action. The most frequent class that dropped its performance was "BreastStroke" for M-MC3, "Shotput" in both M-R(2+1)D, and M-R3D based variants.

Among all the testing designs, only the based on M-R(2+1)D, variant 3-Temporal, had the highest minimum accuracy recorded with a 51.43% detected on "Nunchucks". This case was the only exception among the 33 tests to pass a value of 50 % accuracy, as the rest of the models failed to reach.

We further widened our tests to include the worst accurate classes for the original M-MC3, M-R3D, and M-R(2+1)D models and then studied the effect of such variation made in the variants on these actions. Table 4 – 6 list the five actions that were the most misclassified and the value of their accuracies per variant.

**Table 4.** Accuracy improvement associated with variants on original M-MC3 most five misclassified actions

| Least accurate classes | M-MC3 | FC-Spatial | FC-Temporal | 3-SE | 3-Temporal | 3-Both | 3-CBAM | 3-TCN | All-SE | All-Temporal | All-Together |
|---|---|---|---|---|---|---|---|---|---|---|---|
| BrushingTeeth | 36.11 | 52.78 | 50 | 55.56 | 61.11 | 44.44 | 52.78 | 47.22 | 44.44 | 47.22 | 50 |
| WalkingWithDog | 33.33 | 52.78 | 58.33 | 75 | 77.78 | 63.89 | 52.78 | 75 | 52.78 | 63.89 | 66.67 |
| HandstandWalking | 26.47 | 52.94 | 50 | 50 | 44.12 | 38.24 | 50 | 35.29 | 47.06 | 35.29 | 38.24 |
| PizzaTossing | 24.24 | 33.33 | 30.3 | 30.3 | 48.48 | 45.45 | 45.45 | 60.61 | 30.3 | 33.33 | 54.55 |
| HighJump | 19.92 | 29.73 | 24.32 | 37.84 | 40.54 | 51.35 | 45.95 | 45.95 | 51.35 | 54.05 | 43.24 |

In Table 4, the activity "BrushingTeeth" had the most increase in accuracy by the 3–Temporal model, with added 25% extra value to its original value achieved by M-MC3. The same variant among all ten, was also the most booster for the "WalkingWithDog" action, which was originally at 33.33 to become 77.78 %.

The original accuracy of 26.47% for class "HandstandWalking" was doubled in the FC-Spatial test case, followed by a value of 50% in both FC-Temporal and 3-SE. In the case of the "PizzaTossing" class, only the 3-TCN model managed to increase its accuracy above 50% with a value of 60.61%.

For the lowest accurate class in the M-MC3 original test," HighJump", the variant All-Temporal achieved the greatest improvement by adding an extra 34.13% accuracy score to it.

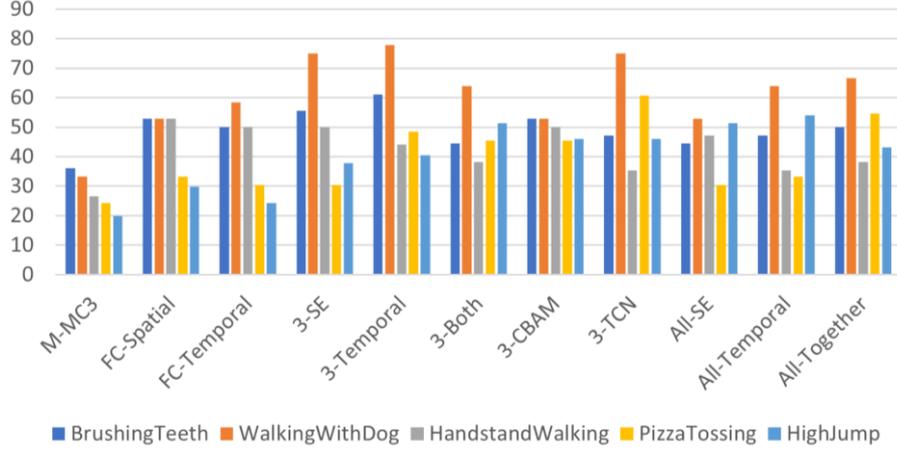

**Fig. 4.** Accuracies of the most five misclassified actions of M-MC3 per its variants.

Figure 4 visualizes the most five misclassified actions tested by M-MC3 and how their accuracies improve over the 10 variants. In Table 5, the case of the five most misclassified actions for M-R(2+1)D shows three common actions as the ones from M-MC3, namely, "BrushingTeeth", "WalkingWithDog", and "PizzaTossing" despite their order.

**Table 5.** Accuracy improvement associated with variants on original M-R(2+1)D most five misclassified actions

| Least accurate classes | M-R(2+1)D | FC-Spatial | FC-Temporal | 3-SE | 3-Temporal | 3-Both | 3-CBAM | 3-TCN | All-SE | All-Temporal | All-Together |
|---|---|---|---|---|---|---|---|---|---|---|---|
| WalkingWithDog | 47.22 | 80.56 | 83.33 | 72.22 | 80.56 | 80.56 | 80.56 | 80.56 | 77.78 | 80.56 | 77.78 |
| BrushingTeeth | 44.44 | 47.22 | 50 | 36.11 | 61.11 | 61.11 | 58.33 | 55.56 | 47.22 | 44.44 | 52.78 |
| HighJump | 37.84 | 51.35 | 59.46 | 54.05 | 67.57 | 54.05 | 51.35 | 62.16 | 64.86 | 67.57 | 81.08 |
| BreastStroke | 35.71 | 92.86 | 85.71 | 85.71 | 60.71 | 85.71 | 57.14 | 64.29 | 75 | 92.86 | 67.86 |
| PizzaTossing | 27.27 | 42.42 | 57.58 | 45.45 | 54.55 | 39.39 | 42.42 | 69.7 | 30.3 | 57.58 | 63.64 |

Table 5 indicates a total increase of 36% was added to the original 47.22% accuracy by the test on variant FC-temporal. Moreover, all variants pushed the accuracy to more than 77% on this activity. The case is different for "BrushingTeeth" which had an accuracy of 44.44% on the base model, as the 3-Spatial model dropped this value by 8.33%, but both the 3-Temporal and 3-Both succeeded in increasing it to become 61.11%.

The third class in this list was "HighJump", with an original test accuracy of 37.84%, which was more than doubled in the All-Together model. The action by the name "BreastStroke" had the highest increment to its original performance by models FC-Spatial and All-temporal in which an accuracy of 92.86% was achieved by both.

In the case of the "PizzaTossing" action, its accuracy has shifted from its original value of 27.27 % to become 69.7% when tested on the 3-TCN variant.

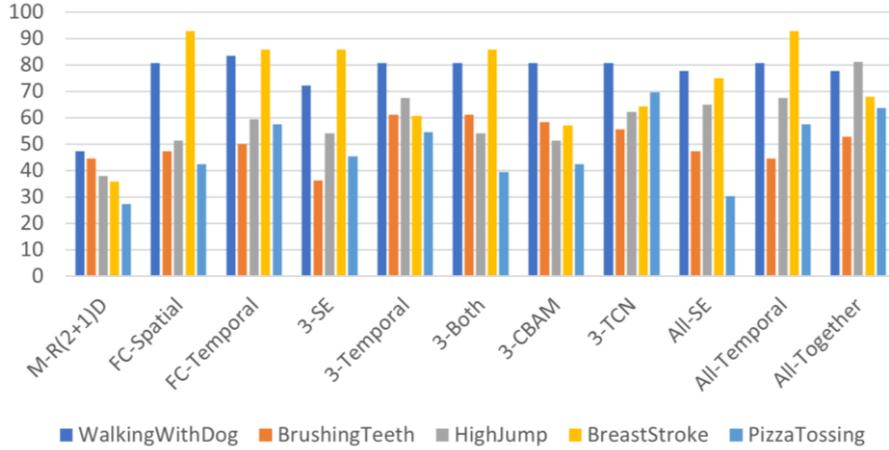

**Fig. 5.** Accuracies of the most five misclassified actions of M-R(2+1)D per its variants.

Figure 5 draws these five actions and their corresponding accuracies across the M-R(2+1)D variants. Table 6 details the performance across M-R3D variants for the classes "Nunchucks", Hammering, "JavelinThrow", "BrushingTeeth", and "PizzaTossing".

**Table 6.** Accuracy improvement associated with variants on original M-R3D most five misclassified actions

| Least accurate classes | M-R3D | FC-Spatial | FC-Temporal | 3-SE | 3-Temporal | 3-Both | 3-CBAM | 3-TCN | All-SE | All-Temporal | All-Together |
|---|---|---|---|---|---|---|---|---|---|---|---|
| Nunchucks | 40 | 51.43 | 57.14 | 42.86 | 57.14 | 25.71 | 34.29 | 42.86 | 37.14 | 42.86 | 42.86 |
| Hammering | 39.39 | 63.64 | 48.48 | 57.85 | 48.48 | 63.64 | 48.48 | 39.39 | 45.45 | 40.48 | 48.48 |
| JavelinThrow | 35.48 | 61.29 | 64.52 | 70.97 | 64.52 | 51.61 | 74.19 | 61.29 | 70.97 | 70.97 | 74.19 |
| BrushingTeeth | 19.44 | 55.56 | 61.11 | 44.44 | 61.11 | 50 | 52.78 | 55.56 | 50 | 47.22 | 36.11 |
| PizzaTossing | 9.09 | 21.21 | 30.3 | 15.15 | 30.3 | 21.21 | 30.3 | 12.12 | 27.27 | 33.33 | 18.18 |

The results shown in Table 6 indicate that the first class had a maximum increase of 17.14% by 3-Temporal. However, the "Hammering" class was nearly doubled in accuracy in the FC-Spatial setup. The "JavelinThrow" action had the highest increase to its original accuracy by the 3-CBAM model recorded 74.19% compared to the original 35.48%. Model All-Together, as Table 6 shows, was the least class that improved among the five considered ones, "BrushingTeeth", but the two models of 3-Temporal and FC-Temporal were beneficial to this action by enhancing its original value from 19.44% to become slightly above 61%. The most misclassified class in the M-R3D test was "PizzaTossing", with an accuracy of only 9.09%, which was never increased to more than 33.33% by the All-Temporal variant.

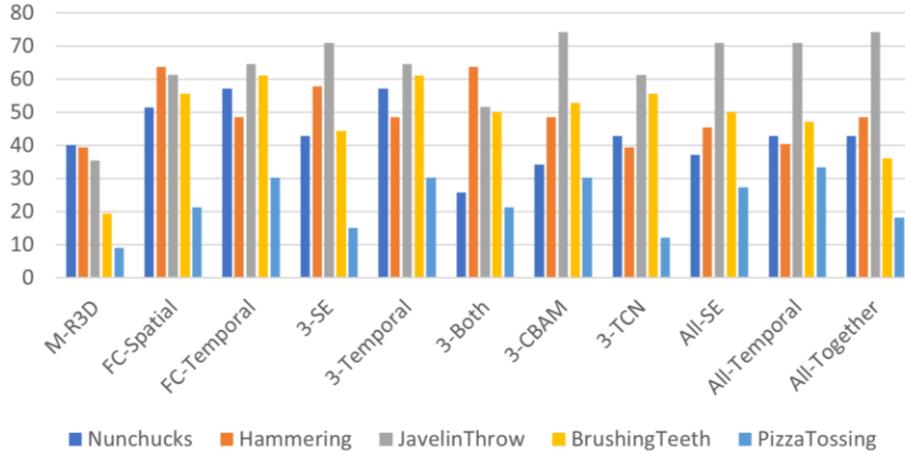

**Fig. 6.** Accuracies of the most five misclassified actions of M-R3D per its variants.

Figure 6 visualizes the least accurate five actions of M-R3D variants. Tables 4-6 show that there are common actions among the three backbones having the lowest accuracies, i.e. "PizzaTossing", and "BrushingTeeth". Generally, the lowest accurate activities in M-MC3, M-R(2+1)D, and M-R3D were improved by their variants. Some of these low-performance actions were positively affected by special attention types implemented in a corresponding variant, as the case of "BreastStroke" in M-R(2+1)D, and its variant FC-Spatial that improved the original accuracy by the backbone from 35.71 to 92.86%. We also observed that three variants only had a slight to moderate enhancement, i.e. "PizzaTossing".

Our work in this paper was carried out for the purpose of indicating the importance of temporal information that should be passed on to a 3D CNN based model. For this purpose, we referred to the models MC3, R3D, and R(2+1)D, which are similar-in architectures state-of-the-art models, but different in their method of extracting the spatial and temporal features due to how their filters are presented in their CNN, for example, if we look at the filters in layer 3 in each of these three models, we see that R3D uses (3, 3, 3), and (1, 3, 3) in MC3, and both of (1, 3, 3) and (3, 1, 1) in the case of R(2+1)D.

In our experiments, we initially set up a mechanism that allowed a strict extent of motion to be gathered from videos, which we believed would hinder the model's learning capabilities. We then increased the spatial features in our data augmentation to observe if this step could compensate for the loss of temporal data. Our findings indicated that enhanced appearance could not substitute for that loss in temporal dynamics.

According to the three original models in the paper [25], which stated only R(2+1)D to be tested on UCF101 and having an accuracy of 96.8%, where our corresponding M-R(2+1)D scored an accuracy of 85.14% on the same dataset. Several factors contributed to this performance gap. Firstly, our testing procedure only sampled three random clips per video, which provided insufficient temporal coverage. Secondly, we did not include spatial augmentations in the test transformation, as in training augmentation. Additionally, the randomness in the starting points of the sampled clips further introduced variance, potentially missing critical temporal activity segments. Moreover, our training mechanism halted the process based on early stopping, as indicated in Tables 1, 2, and 3, compared to the longer training epochs of the original work. Lastly, the added dropout layer before the fc one might hinder optimal learning, despite its advantage in generalizing the model.

In the second part of our experiment, we developed 10 new variants for each backbone model. Each of them employs a different attention mechanism block in one or more locations within the CNN layers in the backbones. Our motivation behind that was, first, to study both the effect and the extent of improvement that such attention block would bring to the temporal-restricted models we earlier created, and second, the actions that this added attention could strongly influence. Our results show that these blocks behave similarly for all base models. As for instance, what a temporal attention block on a variant based on R3D enhanced, was not the same case as in the MC3 setup.

In comparison to the related research, paper [38] proposed an R(2+1)D model that adopted a resolution of 112 x 112, in addition to a mechanism to drop frames during training, along a temporal decay of 0.999, and achieved an accuracy of 78.7% on UCF101. Another approach [39] was based on the ResNet-154 transformer model, containing 10 frames, that did lead to an accuracy of 76.64% when testing on UCF101. Utilizing 3D CNN in HAR was also proposed by [40], authors of this work have tested their model on different datasets such as UCF50 and UCF101, on which they scored a value of 79.9% accuracy on the latter dataset. They also stated that the class "Horse-Riding" was the most misclassified one. On the other hand, among all our tests, the best performance was related to the variant 3-Temporal of M-R(2+1)D model, which achieved an accuracy of 88.98%. "Nunchuks" action was the most misclassified activity related to this model, with a value of 51.43% accuracy.

## 5     Conclusions

In this paper, we carried out a comparative analysis among three 3D CNN Resnet-based architectures, namely the MC3, R(2+1)D, and R3D models, by following a mechanism that restricts the knowledge gathered from motion in video frames; we also increased the resolution of the frames to be double compared to how the training was set on the three original models. We created similar designs that differ in adding a dropout layer for better generalization and trained them on the UCF101 dataset. At another phase of the experiment, we referred to our three modified models named M-MC3, M-R(2+1)D, and M-R3D as backbones. Our results indicated an overall drop in modes performance compared to the original work and the backbones showed accuracies of 81.21%, 85.14 %, and 79.99 for M-MC3, M-R(2+1)D, and M-R3D, respectively. This implied the importance of the temporal features in all the 3D CNN models despite their used filters. In addition to that, another observation we noted was that the enhanced resolution in frames could not compensate for the effect of missing skipped frames.

Based on our modified models, we then developed 10 variants for each, that are similar to the backbone, except that they employ different blocks that utilize attention mechanisms to operate on temporal data, and channel levels. Testing a total of 30 models has shown an increase of up to 4% to the backbones' accuracies. Furthermore, as these variants are relatively close in their overall effect on the models' performance, they differ in their behavior, i.e. their class accuracy levels.

In our future work, we would like to study these added units on different architectures rather than 3D-based models. For instance, on video Transformer, two-streams models will be used, so that we can analytically compare their results with 3D CNNs.

**Disclosure of Interests.** The authors have no competing interests to declare that are relevant to the content of this article.